\documentclass[12pt]{spieman}  
\usepackage{amsmath,amsfonts,amssymb}
\usepackage{graphicx}
\usepackage{setspace}
\usepackage{tocloft}
\usepackage{graphicx}
\usepackage{textcomp}
\usepackage{hhline}
\usepackage{adjustbox}
\usepackage{cite}
\usepackage[numbers,sort&compress]{natbib}
\usepackage{amssymb}
\usepackage{amsfonts}
\usepackage[utf8]{inputenc}
\usepackage[T1]{fontenc} 
\usepackage{enumerate}
\usepackage{booktabs}
\usepackage{algorithm,algpseudocode}

\usepackage{color}
\usepackage{setspace}
\usepackage{url}
\usepackage{bm}
\usepackage{ctable}
\usepackage{multirow}
\usepackage{multicol}
\usepackage{caption}

\usepackage{lineno}

\title{Image inpainting using frequency domain priors}

\author[a,*]{Hiya Roy}
\author[b]{Subhajit Chaudhury}
\author[b]{Toshihiko Yamasaki}
\author[a]{Tatsuaki Hashimoto}
\affil[a]{The University of Tokyo, Department of Electrical Engineering and Information Systems, 7-3-1 Hongo, Bunkyo-ku, Tokyo, Japan, 113-8656.}
\affil[b]{The University of Tokyo, Department of Information and Communication Engineering, 7-3-1 Hongo, Bunkyo-ku, Tokyo, Japan, 113-8656.}

\cftpagenumbersoff{figure}
\cftpagenumbersoff{table} 

\begin{document} 
	\newcommand{\norm}[1]{\lVert#1\rVert}
\maketitle
%\linenumbers
\begin{abstract}
In this paper, we present a novel image inpainting technique using frequency domain information. Prior works on image inpainting predict the missing pixels by training neural networks using only the spatial domain information. However, these methods still struggle to reconstruct high-frequency details for real complex scenes, leading to a discrepancy in color, boundary artifacts, distorted patterns, and blurry textures. To alleviate these problems, we investigate if it is possible to obtain better performance by training the networks using frequency domain information (Discrete Fourier Transform) along with the spatial domain information. To this end, we propose a frequency-based deconvolution module that enables the network to learn the global context while selectively reconstructing the high-frequency components. We evaluate our proposed method on the publicly available datasets CelebA, Paris Streetview, and DTD texture dataset, and show that our method outperforms current state-of-the-art image inpainting techniques both qualitatively and quantitatively. 
\end{abstract}


{\noindent \footnotesize\textbf{*}Hiya Roy,  \linkable{hiya@hal.t.u-tokyo.ac.jp} }

\section{Introduction}\label{intro}  
\indent In computer vision, the task of filling in missing pixels of an image is known as image inpainting. It can be extensively applied for creative editing tasks such as removing unwanted/distracting objects in an image or generating the missing region of an occluded image or improving data availability for satellite imagery. The main challenge in this task is to synthesize the missing pixels in such a way that it looks visually realistic and coherent to human eyes. 

\indent Traditional image inpainting algorithms~\cite{bertalmio2000image, ballester2001filling, bertalmio2003simultaneous, levin2003learning, efros2001image, drori2003fragment, criminisi2004region, simakov2008summarizing, barnes2009patchmatch, darabi2012image, he2014image} can be broadly divided into two categories. Diffusion-based image inpainting algorithms~\cite{bertalmio2000image, ballester2001filling, bertalmio2003simultaneous, levin2003learning} focus on propagating the local image appearance into the missing regions. Although these methods can fill in small holes but produce smoothed results as the hole grows bigger. On the other hand, patch-based traditional inpainting algorithms~\cite{efros2001image, drori2003fragment, criminisi2004region, simakov2008summarizing, barnes2009patchmatch, darabi2012image, he2014image} iteratively search for the best-fitting patch in the image to fill in the missing region. These methods can fill in bigger holes, but they are not effective either in inpainting missing regions that have complex structures or in generating unique patterns or novel objects that are not available in the image in the form of a patch.

\indent Recent research works on image inpainting~\cite{pathak2016context, iizuka2017globally, yu2018generative, song2018spg, nazeri2019edgeconnect, yu2019free} leverage the advancements in generative models such as Generative Adversarial Networks~(GANs)~\cite{goodfellow2014generative} and show that it is possible to learn and predict missing pixels in coherence with the existing neighboring pixels by training a convolutional encoder-decoder network. In this paradigm, generally speaking, the model is trained in a two-stage manner - i) in the first stage, the missing regions are coarsely filled in with initial structures by minimizing traditional reconstruction loss; ii) in the second stage, the initially reconstructed regions are refined using an adversarial loss. Although these methods are good in generating visually plausible novel contents such as human faces, structures, natural scenes in the missing region, they still struggle to reconstruct high-frequency details for real complex scenes, leading to a discrepancy in color, boundary artifacts, distorted patterns, and blurry textures. Additionally, the reconstruction quality of previous methods deteriorates as the size of the missing region increases. The above problems can be attributed to the following reason. Existing methods use only spatial domain information during the learning process similar to diffusion like techniques to obtain information from the mask boundary. Thus as the mask size increases, the interior reconstruction details are lost and only a low-frequency component of the original patch is estimated by these methods.  

\begin{figure*}[tb]
	\centering
	\includegraphics[width=0.9\linewidth]{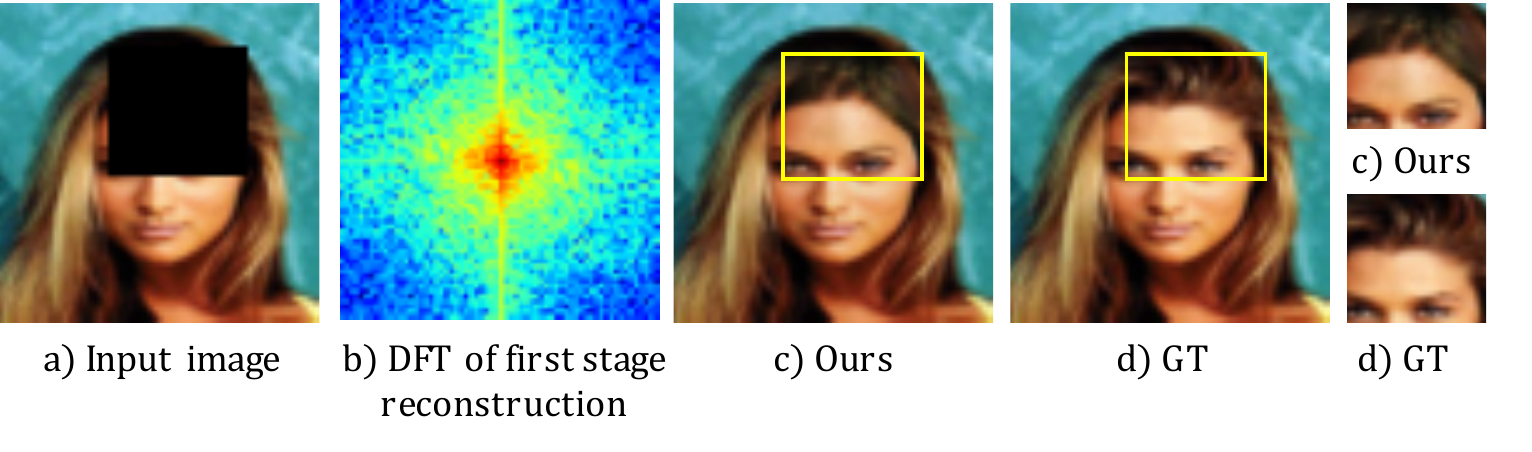}
	\vspace{2mm}
	\caption{a) Input images with missing regions, b) DFT of first stage reconstruction by our deconvolution network, c) Image inpainting results (after the second stage) of our proposed approach, and d) Ground Truth (GT) image. The last column shows the prediction of the missing region obtained from our method and original pixel values for the same region in the GT image.}\label{fig:first}
\end{figure*}

\indent To alleviate the above problem, we resort to frequency-based image inpainting. We show that image inpainting can be converted to the problem of deconvolution in the frequency domain which can predict local structure in the missing regions using global context from the image. Qualitative analysis shows that our proposed frequency domain image inpainting approach helps in improving the texture details of missing regions. Previous methods make use of only spatial domain information. Therefore, the reconstruction of the information close to the mask boundary is good compared to the interior region since the local context is available only in the boundary regions. In contrast, a frequency-based approach would take information from the global context in the image because to Discrete Fourier Transforms (DFT) that considers all pixels for computing the frequency components. As a result, it captures more detailed structural and textural content of the missing regions in the learned representation. Due to these reasons, we propose a two-stage network consisting of i) deconvolution stage and ii) refinement stage. In the first stage, the DFT image from the original RGB image is computed. Each frequency component in the DFT image captures the global context thus forming a better representation of the global structure. We employ a Convolutional Neural Network (CNN) to learn the mapping between masked DFT and original DFT, which is a deconvolution operation obtained by minimizing the $\ell_2$ loss. While DFT based deconvolution can reconstruct the global structural outline, we observe that there exists a mismatch in the color space of the first stage output. Therefore, in the second stage, we fine-tune the output of the first stage using adversarial methods to match the color distribution of the true image. Figure~\ref{fig:first} shows an example of the reconstructed output using our method where Figure~\ref{fig:first}b) shows the DFT map of our first stage reconstruction obtained from the deconvolution network). This additional frequency domain information is later used by the refinement network to obtain the final output as shown in Figure~\ref{fig:first}c).
Our main contributions in this paper can be summarized as follows:
\begin{enumerate}
	\item We introduce a new frequency domain-based image inpainting framework that learns the high-frequency component of the masked region by using the global context of the image. We find that the network learns to preserve image information in a better way when it is trained in the frequency domain. 
	Therefore, adding the frequency domain and spatial domain information certainly improves the inpainting performance compared to the conventional spatial domain image inpainting algorithms. To enable better inpainting, we train the network using both frequency-domain and spatial domain information which leads to a better consistency of inpainted results in terms of the local and global context. 
	\item We validate our method on benchmark datasets including CelebA faces, Paris Streetview, and DTD texture datasets, and show that our method achieves better inpainting results in terms of visual quality and evaluation metrics outperforming the state-of-the-art results. To the best of our knowledge, this is the first work that explores the benefits of using frequency domain information for image inpainting.
\end{enumerate}

\section{Related Work}
\subsection{Traditional Inpainting Techniques} 
\textbf{Diffusion-based} image completion methods~\cite{bertalmio2000image, ballester2001filling, bertalmio2003simultaneous, levin2003learning} are based on Partial Differential Equations (PDE) where a diffusive process is modeled using PDE to propagate colors into the missing regions. These methods work well for inpainting small missing regions but fail to reconstruct the structural component or texture for larger missing regions.\\
\textbf{Patch-based} algorithms, on the other hand, are based on iteratively searching for similar patches in the existing image and paste/stitch the most similar block onto the image. Efros and Freeman~\cite{efros2001image} first proposed a patch-based algorithm for texture synthesis based on this philosophy. These algorithms perform well on textured images by assuming that the texture of the missing region is similar to the rest of the image. However, they often fail in inpainting missing regions in natural images because the patterns are locally unique in such images. Moreover, these methods are computationally expensive because of the need for computing similarity scores for every target-source pair. For more accurate and faster image inpainting, several optimal patch search based methods were proposed by Drori et al.~\cite{drori2003fragment} (fragment-based image completion algorithm) and Criminisi et al.~\cite{criminisi2004region} (patch-based greedy sampling algorithm). Another optimization method to synthesize visual data (images or video) based on bi-directional similarity measure was proposed by Simakov et al.~\cite{simakov2008summarizing}. Afterward, these techniques were expedited by Barnes et al.~\cite{barnes2009patchmatch} who proposed PatchMatch, a fast randomized patch search algorithm that could handle the high computational and memory cost. Later such patch-based image completion techniques were improved by Darabi et al.~\cite{darabi2012image} by incorporating gradient-domain image blending, He et al.~\cite{he2014image} by computing the statistics of patch offsets and Ogawa et al.~\cite{ogawa2013image} by optimizing sparse representations w.r.t. SSIM perceptual metric. However, these methods rely only on existing image patches and use low-level image features. Therefore they are not effective in filling complex structures by performing semantically aware patch selections. 

\subsection{Deep Learning-based Inpainting}
\indent Recently CNN models~\cite{lecun1990handwritten} have shown tremendous success in solving high-level tasks such as classification, object detection, and segmentation as well as low-level tasks such as image inpainting problem. Xie et al.~\cite{xie2012image} proposed Stacked Sparse Denoising Auto-encoders~(SSDA), that combines sparse coding and deep networks pre-trained with denoising auto-encoder to solve a blind image inpainting task. Blind image inpainting is harder because in this case, the missing pixel locations are not available to the algorithm and it has to learn to find the missing pixel location and then restore them. Kohler et al.~\cite{kohler2014mask} showed a mask specific deep neural network-based blind inpainting technique for filling in small missing regions in an image. Chaudhury et al.~\cite{chaudhury2017can} attempted to solve this problem by proposing a lightweight fully convolutional network~(FCN) and demonstrated that their method can achieve comparable performance as the sparse coding based $K$-singular value decomposition (K-SVD)~\cite{mairal2007sparse} technique. However, these inpainting approaches were limited to very small sized masks.

\indent More recently, adversarial learning-based inpainting algorithms have shown promising results in solving image inpainting problems because of their ability to learn and synthesize novel and visually plausible contents for different images such as objects~\cite{pathak2016context}, scene completion~\cite{iizuka2017globally}, faces~\cite{yeh2017semantic} etc. A seminal work by Pathak et al.~\cite{pathak2016context} showed that their proposed Context Encoder network can predict missing pixels of an image based on the context of the surrounding areas of that region. They used both standard \(\ell_2\) loss and adversarial loss~\cite{goodfellow2014generative} to train their network. Later, Iizuka et al.~\cite{iizuka2017globally} demonstrated that their encoder-decoder model could reconstruct pixels in the missing region that are consistent both locally and globally, by leveraging the benefits of dilated convolution layers, a variant of standard convolutional layers. Similar to \cite{pathak2016context} this approach also uses adversarial learning for image completion, but unlike \cite{pathak2016context} it could handle arbitrary image and mask size, because of the proposed global and local context discriminator networks. Recently, Yu et al.~\cite{yu2018generative} introduced the concept of attention for solving an image inpainting task by proposing a novel contextual attention layer and trained the unified feedforward generative network with reconstruction loss and two Wasserstein GAN losses~\cite{arjovsky2017wasserstein, gulrajani2017improved}. They showed that their method can inpaint images with multiple missing regions having different sizes and located arbitrarily in the image. Later, Liu et al.~\cite{liu2018image} proposed a partial convolution layer with an automatic mask-update rule, that can handle free-form/irregular masks. Here, the mask is updated in such a way that the missing pixels are predicted based on the real pixel values of the original image where the partial convolution can operate. Song et al.~\cite{song2018spg} showed that it is possible to perform image inpainting by using segmentation information. To this end, they proposed a model that predicts the segmentation labels of the corrupted image at first and then fills in the segmentation mask so that it can be used as guidance to complete the image. Nazeri et al.~\cite{nazeri2019edgeconnect} introduced an edge generator that at first predicts the edges of the missing regions and then use the predicted edges as a guidance to the complete the image. Yu et al.~\cite{yu2019free} proposed a gated convolution based approach to handle free-form image completion.

\begin{figure*}[tb]
	\centering
	\includegraphics[width=\linewidth]{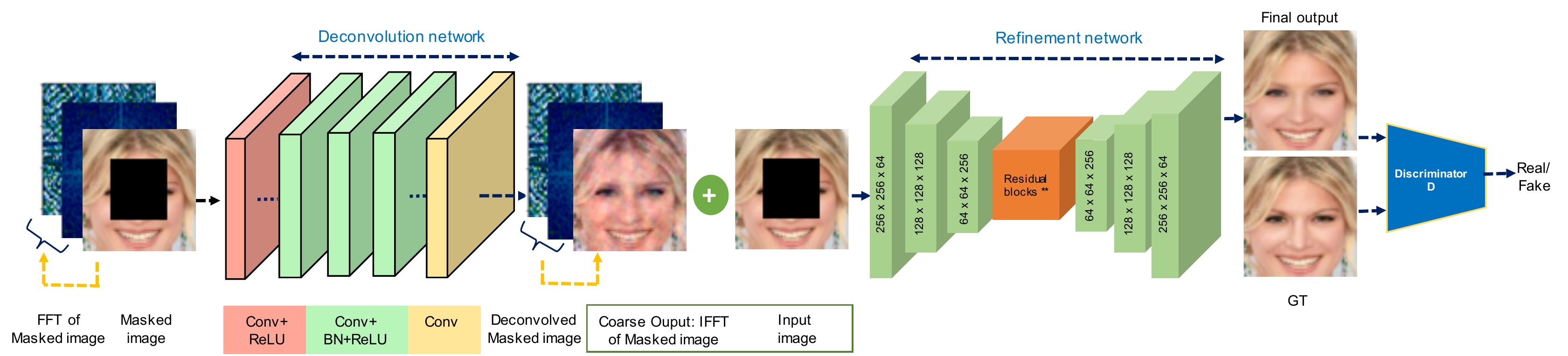}
	\vspace{1mm}
	\caption{Overview of our frequency domain-based image inpainting framework. The deconvolution network is trained in the frequency domain with $\ell_2$ loss to learn the mapping between DFT of masked image and the original image. The refinement network is trained in the spatial domain with adversarial loss.}\label{fig:proposed}
\end{figure*}

\subsection{Frequency-domain Learning}
Recently enabling the network to learn information in the frequency domain has gained popularity because the frequency domain information contains rich representations that allow the network to perform the image understanding tasks in a better way in comparison to the conventional way of using only spatial domain information. Gueguen et al.~\cite{gueguen2018faster} proposed image classification using features from the frequency domain. Xu et al.~\cite{xu2020learning} showed that it is possible to perform object detection and instance segmentation by learning information in the frequency domain with a slight modification to the existing CNN models that use RGB input. In this paper, we propose to use frequency-domain information along with spatial domain information to achieve better image inpainting performance.

\section{Proposed Method} \label{our_method}
Given a corrupted input image, our aim is to predict the missing region in such a way that it looks similar to the clean images to human eyes. In this paper, we propose a frequency-based non-blind image inpainting framework that consists of two stages: i) frequency domain deconvolution network and ii) refinement network. The overall framework of the proposed method is shown in Figure~\ref{fig:proposed}. In the first stage, we compute the DFT of the masked image~(both magnitude and phase information) and the original RGB image and train a CNN for deconvolution to learn the mapping between the two signals by minimizing the $\ell_2$ loss. Here we formalize the problem of inpainting in the spatial domain as deconvolution in the frequency domain. We employ the feed-forward denoising convolutional neural networks (DnCNNs)~\cite{zhang2017beyond}, a manifestation of deconvolution, which uses residual learning to predict the denoised image. The motivation behind this DFT-based deconvolution operation is to learn a better representation of the global structure that can serve as guidance to the second network.
In the second stage, we use the spatial domain information (of the masked image and the mask) and train a generative adversarial network~(GAN) based model~\cite{goodfellow2014generative} by minimizing an adversarial loss along with $\ell_2$ loss. The motivation to incorporate this stage is to fine-tune the output of the first stage by refining the structural details and matching the color distribution of the true image in a local scale. The various components of our model are explained in the following subsections.

\begin{figure*}[tb]
	\centering
	\includegraphics[width=0.9\linewidth]{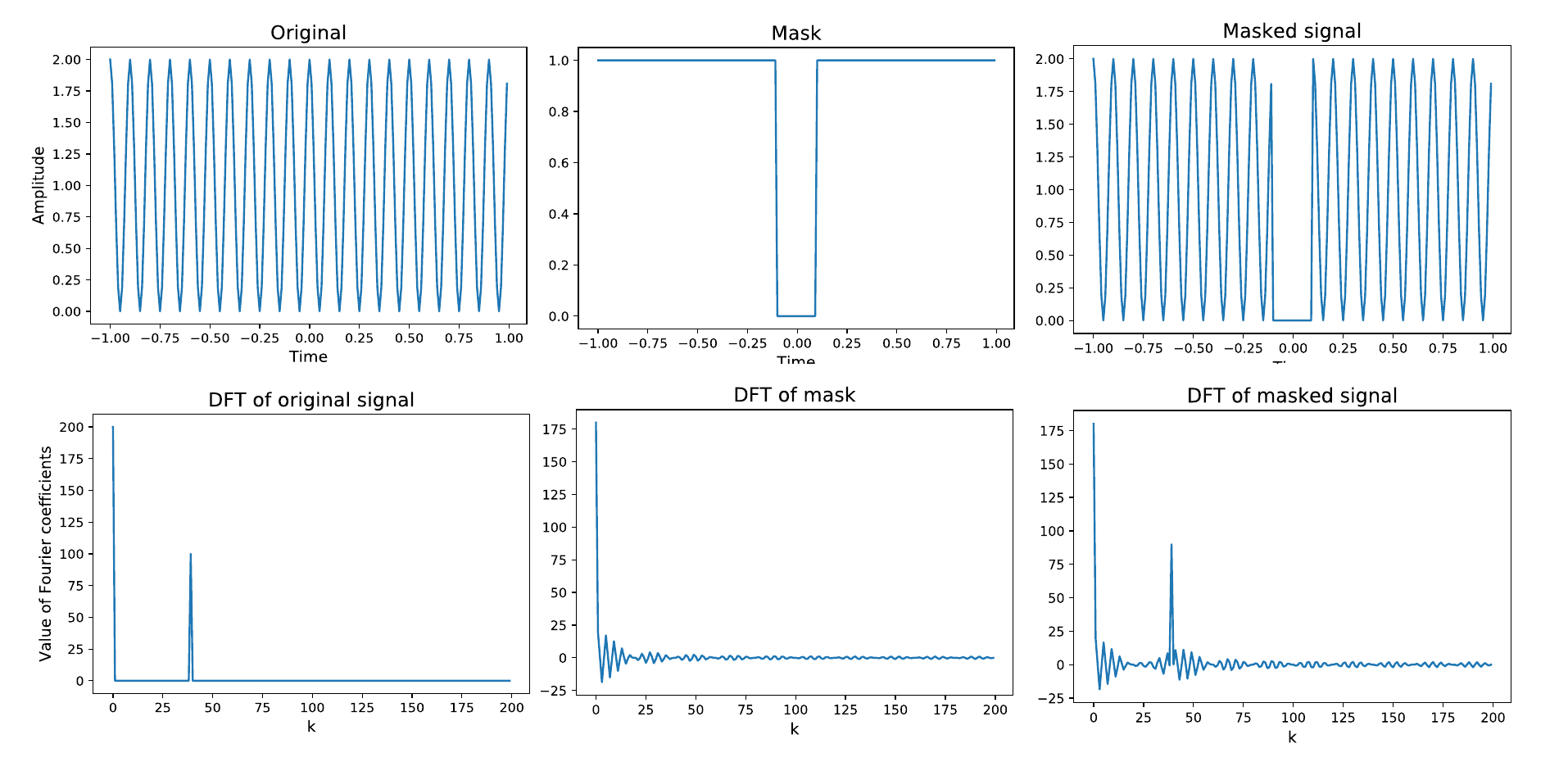}
	\caption{Visualization of masked signal in frequency domain~(using DFT). Here, we use the convolution-multiplication property of DFT to transform signals from spatial to frequency domain and vice-versa.}\label{fig:dft}
\end{figure*}

\subsection{Frequency-domain Deconvolution Network}\label{1ststage}
\subsubsection{Problem Formulation} 
Let us consider \(\mathbf{I_{in}}\) as the corrupted/incomplete input image, \(\mathbf{I_{gt}}\) as the ground truth image, and \(\mathbf{I^1_{pred}}\) as the predicted output image after the first stage. At first, we calculate the DFT of \(\mathbf{I_{in}}\) and \(\mathbf{I_{gt}}\) as \(\mathbf{I^f_{in}}\) = DFT(\(\mathbf{I_{in}}\)) and \(\mathbf{I^f_{gt}}\) = DFT(\(\mathbf{I_{gt}}\)). Let us consider a mask function in spatial domain as \(\mathbf{M}\), with its frequency domain counterpart as \(\mathbf{M}^f\). 

\noindent A masked image is represented as $\mathbf{I_{in}}(x,y) = \mathbf{I_{gt}}(x,y) \odot \mathbf{M}(x,y)$ where $\odot$ denotes element-wise multiplication. Our contribution in this paper is to analyze this relation between the frequency domain signals of \(\mathbf{I_{in}}\), \(\mathbf{I_{gt}}\), and \(\mathbf{M}\). For example, if we consider a mask of size $(2W,2H)$, the power spectral density for the DFT of mask signal can be given as

%\begin{center}
\begin{align}
|\mathbf{M}^f(p,q)|^{2} \propto \frac{sin({\pi p})}{sin(\frac{\pi p}{N})}\frac{sin({\pi q})}{sin(\frac{\pi q}{N})}, \label{dft_equation}
\end{align}
\noindent where $k = 0, 1, ... (N-1)$ represents the discrete frequency, with $N$ being the number of samples. The frequency domain representation of the mask signal is shown in Figure~\ref{fig:dft}, which depicts a decaying pulse from the origin. By the convolution-multiplication property of DFT, we can show that the multiplication of mask with the image in spatial domain is equivalent to convolution of mask with image in frequency domain (Figure~\ref{fig:dft}). Mathematically, this is represented as
\begin{equation}
\mathbf{I^f_{in}}(p,q)=\mathbf{I^f_{gt}}(p,q) \circledast \mathbf{M}^f(p,q) \label{conv}
\end{equation}
\noindent where $\circledast$ denotes the convolution operation and the masked frequency signal is the output of the convolution of the mask and clean image DFT signal. Therefore, we perform a deconvolution operation to predict the missing region of the incomplete image. Let \(\mathbf{F}(\mathbf{I_{in};\theta})\) be the Deconvolutional neural network that converts \(\mathbf{I_{in}}\) to \(\mathbf{I^1_{pred}}\), such that \(\mathbf{I^1_{pred}}\) = \(\mathbf{F}(\mathbf{I_{in};\theta})\). 
After calculating \(\mathbf{I^f_{in}}\) and \(\mathbf{I^f_{gt}}\), we train the network to learn the mapping between them, to predict the first stage output. We denote frequency domain representation as \(\mathbf{I^{1f}_{pred}}\) where  \(\mathbf{I^{1f}_{pred}}\) = \(\mathbf{F}(\mathbf{I^f_{in};\theta})\). Next, we perform an inverse DFT of the first stage output and get the predicted output image \(\mathbf{I^1_{pred}}\) = IDFT(\(\mathbf{I^{1f}_{pred}}\)).

\subsubsection{Network Architecture} 
%Add a table 
To perform the deconvolution operation in the frequency domain, we adopt a CNN model having 17 layers similar to \cite{zhang2017beyond}. This deconvolution network contains three types of layers as shown in Figure~\ref{fig:proposed}. The first layer is a Conv layer with ReLU non-linearity where 64 filters of (3x3x3) size are used. Next layers~($2^{nd}$-$16^{th}$) are a combination of Conv layer, a batch normalization layer~\cite{ioffe2015batch} and a ReLU layer, where 64 filters of (3×3×64) size are used. The last layer is a Conv layer, where 3 filters of (3x3x64) size are used to reconstruct the output. Details of our first stage deconvolution network is given in Table~\ref{deconvolution_net}.

	\begin{table}[t]
	\centering
	\caption{First stage network architecture (Deconvolution network).}
	\label{deconvolution_net}
	\begin{adjustbox}{width=0.9\textwidth}
		\def\arraystretch{1.2}
		\begin{tabular}{ccccc}
			\hline
			\specialrule{.1em}{.1em}{.1em} 
			 \textbf{Layer name}  \quad & \quad \textbf{Layer no.}  \quad & \quad \textbf{Stride, Padding}  & \quad  \textbf{Activation} & \quad \textbf{Layer output size} \\ \hline
			\specialrule{.1em}{.1em}{.1em} 
			Input  &  \quad - \quad & \quad - \quad & \quad - \quad & \quad $1\times12\times64\times64$ \\ \hline
			Conv $3\times3$ \quad   & \quad 1 \quad & \quad 1, 1 \quad & \quad ReLU   \quad  &  \quad \\ \hline
			Conv $3\times3$   & \quad 2-16 (15 layers) \quad & \quad 1, 1 \quad & \quad (Batch Norm+ReLU)  \quad &      \\ \hline
			Conv $3\times3$   & \quad 17 \quad & \quad - \quad & \quad - \quad & \quad $1\times6\times64\times64$ \\ \hline
			\specialrule{.1em}{.1em}{.1em} 
		\end{tabular}
	\end{adjustbox}
\end{table}

\subsubsection{Training} 
To train our deconvolution network, we use \(\ell_2\) loss that minimizes the distance between the DFT of ground-truth image \(\mathbf{I^f_{gt}}\) and the DFT of inpainted image \(\mathbf{I^{1f}_{pred}}\), which is given by
\begin{align}
\mathcal{L}_{s1} = \left\|\mathbf{I^f_{gt}} - \mathbf{I^{1f}_{pred}}\right\|_{2}^{2}
\end{align}
After training the first stage deconvolution network, we compute the inverse DFT of \(\mathbf{I^{1f}_{pred}}\) which is used as a guidance to train the refinement stage as shown in Figure~\ref{fig:proposed}. The reason to choose the frequency domain in the first network is because it contains rich information~\cite{xu2020learning, xu2018lapran} for high frequency preservation. 

\subsection{Refinement Network}\label{inpainting_module}
\indent The refinement network is a GAN based model~\cite{goodfellow2014generative} that has shown promising results in generative modeling of images~\cite{radford2015unsupervised} in recent years. Our refinement network has a generator and a discriminator network, where the generator network takes the output of the first stage (frequency domain deconvolution module), the original masked image, and the corresponding binary mask (spatial domain information) as input pairs, and outputs the generated image. The discriminator network takes this generator output and minimizes the Jensen–Shannon divergence between the input and output data distribution to match the color distribution and structural details of the output image to the true image. 

\subsubsection{Network Architecture}
\textbf{Generator:} We adapt the generator architecture from Johnson et al.~\cite{johnson2016perceptual} that has exhibited good performance for image-to-image translation task~\cite{zhu2017unpaired}. Our generator network is an encoder-decoder architecture having three convolution layers for downsampling, eight residual blocks \cite{he2016deep}, and three convolution layers for up-sampling. Here, Conv-2 and Conv-3 layers are stride-2 convolution layers that are responsible for down-sampling twice, and Conv-4 and Conv-5 layers are transpose convolution layers that are responsible for up-sampling twice back to the original image size. We use instance normalization~\cite{ulyanov2016instance} and ReLU activation function across all layers of the generator network. \\

\noindent \textbf{Discriminator: }We adapt the discriminator network from~\cite{isola2017image, zhu2017unpaired} which is a Markovian discriminator similar to 70$\times$70 PatchGAN. The advantage of using a PatchGAN discriminator is that it has fewer parameters compared to a standard discriminator because it works only on a particular image patch instead of an entire image. Furthermore, it can be applied to any arbitrarily-sized images in a fully convolutional fashion~\cite{isola2017image, zhu2017unpaired}. We apply sigmoid function after the last convolution layer which produces a 1-dimensional output score that predicts whether the 70$\times$70 overlapping image patches are real or fake. To stabilize the discriminator network training, we use spectral normalization~\cite{miyato2018spectral} as our weight normalization method. Moreover, we use leaky ReLUs~\cite{maas2013rectifier} with slope of 0.2. The details of our second stage refinement network (generator and discriminator network) and output size of each layer is given in Table~\ref{refinement_net}.

\begin{table}[!htb]
	\caption{Second stage network architecture}
	\label{refinement_net}
	\centering
	\begin{minipage}{.7\linewidth}
		\centering
		\label{generator}
		\begin{adjustbox}{width=0.75\textwidth}
			\def\arraystretch{1.2}
			\begin{tabular}{cccc}
				\hline 
				\multicolumn{4}{c}{\textbf{Generator network}} \\  \hline
				\specialrule{.1em}{.1em}{.1em} 
				\textbf{Layer name} & \textbf{Stride} & \textbf{Activation} & \textbf{Layer output size} \\ 
				\specialrule{.1em}{.1em}{.1em} 
				Input      & -  & - & $1\times9\times64\times64$ \\ \hline
				\multicolumn{4}{c}{\textbf{Encoder network}} \\  \hline
				Conv $7\times7$    & 1 & ReLU  & $1\times64\times64\times64$\\
				Conv $4\times4$    & 2 & ReLU  & $1\times128\times32\times32$ \\ 
				Conv $4\times4$   & 2 & ReLU    & $1\times256\times16\times16$ \\ \hline
				\multicolumn{4}{c}{\textbf{Residual block} ($\times8$)} \\  \hline
				Residual blocks     &          & $1\times256\times16\times16$ \\ \hline
				\multicolumn{4}{c}{\textbf{Decoder network}} \\  \hline
				Conv $4\times4$  & 2 &  ReLU  & $1\times128\times32\times32$ \\
				Conv $4\times4$  & 2 & ReLU      & $1\times64\times64\times64$ \\
				Conv $7\times7$   & 1 &  tanh     & $1\times3\times64\times64$ \\
				\specialrule{.1em}{.1em}{.1em} \\
			\end{tabular}
		\end{adjustbox}
	\end{minipage}%

	\begin{minipage}{.7\linewidth}
		\centering
		\label{discriminator}
		\begin{adjustbox}{width=0.75\textwidth}
		\def\arraystretch{1.2}
		\begin{tabular}{cccc}
			\hline
			\multicolumn{4}{c}{\textbf{Discriminator network}} \\  \hline
			\specialrule{.1em}{.1em}{.1em} 
			\textbf{Layer name}  \qquad &  \qquad \textbf{Stride}  & \qquad  \textbf{Activation} & \qquad \textbf{Layer output size} \\ 
			\specialrule{.1em}{.1em}{.1em} 
			Input      & -  & - & $1\times3\times64\times64$ \\ \hline
			Conv $4\times4$ \qquad   & 2 \qquad & LeakyReLU   \qquad     &  $1\times64\times32\times32$\\ 
			Conv $4\times4$   & 2 & LeakyReLU        &  $1\times128\times16\times16$    \\
			Conv $4\times4$   & 2 & LeakyReLU        & $1\times256\times8\times8$      \\ 
			Conv $4\times4$   & 1 & LeakyReLU        & $1\times512\times7\times7$         \\ 
			Conv $4\times4$   & 1 & Sigmoid             & $1\times1\times6\times6$          \\
			\specialrule{.1em}{.1em}{.1em} 
		\end{tabular}
	\end{adjustbox}
	\end{minipage} 
\end{table}

\begin{algorithm}[t]
	\caption{Training the refinement network.}
	\begin{algorithmic}[1]
		\While {Generator G has not converged}
		\State{Sample batch images \(\mathbf{I_{in}}\) from training data;}
		\State{Generate random masks \(\mathbf{M}\);}
		\State{Construct combined input (\(\mathbf{I_{in}}\), \(\mathbf{M}\), and \(\mathbf{I^1_{pred}}\));}
		\State{Get masked region prediction \(\mathbf{I^2_{pred}} = G(\mathbf{I_{in}},\mathbf{M}, \mathbf{I^1_{pred}})\);}
		\State{Generate inpainted image by modifying the masked region \(\mathbf{I_{pred}} \gets \mathbf{I_{in}}+ \mathbf{I^2_{pred}} \odot (\mathbf{1-M})\);}
		\State{Update G with \(\ell_1\) loss and adversarial critic loss; }
		\State{Update discriminator critic D with \(\mathbf{I_{in}}\), \(\mathbf{I_{pred}}\);}
		\EndWhile
	\end{algorithmic}
	\label{inpainting_algo}
\end{algorithm}

\subsubsection{Training} \label{training}
After obtaining the first stage output, we feed it to the refinement network along with the spatial domain information (of the masked image and the mask). While training, the generator of the inpainting network \(G\) takes a combination of \textit{input image} \(\mathbf{I_{in}}\), \textit{image mask} \(\mathbf{M}\), and the \textit{first stage output image} \(\mathbf{I^1_{pred}}\) and generates \(\mathbf{I^2_{pred}} = G(\mathbf{I_{in}},\mathbf{M}, \mathbf{I^1_{pred}})\) as output. Then by adding \(\mathbf{I^2_{pred}}\) to the \textit{input image}, we get \textit{completed image} as \(\mathbf{I_{pred}} = \mathbf{I_{in}} + [\mathbf{I^2_{pred}} \odot \mathbf{(1-M)}]\). The training procedure of the refinement stage is described in Algorithm \ref{inpainting_algo}. We train our refinement module by using two loss functions: a reconstruction loss and an adversarial loss~\cite{goodfellow2014generative}. 
Here for the reconstruction loss, we use \(\ell_1\) loss~\cite{pathak2016context} that minimizes the distance between the clean/ground-truth image \(\mathbf{I_{gt}}\) and the completed/inpainted image \(\mathbf{I_{pred}}\), which is given by

\begin{align}
\mathcal{L}_{\ell_1}(x) = \norm{\mathbf{I_{gt}} - {\mathbf{I_{pred}}}}_1, \label{L1_equation}
\end{align}
\noindent where $ \mathbf{I_{pred}} \gets \mathbf{I_{in}}+ G(\mathbf{I_{in}},\mathbf{M}, \mathbf{I^1_{pred}}) \odot (\mathbf{1-M})$. For the adversarial loss, we follow the min-max optimization strategy, where the generator $G$ is trained to produce inpainted samples from the artificially corrupted images such that the inpainted samples appear as ``real'' as possible and the adversarially trained discriminator critic $D$ tries to distinguish between the ground truth clean samples and the generator predictions/inpainted samples. The objective function can be expressed as follows

\begin{align}
G^*, D^* = \arg\min_G \max_D \mathcal{L}_{adv}(G,D) = \mathbb{E}_{\mathbf{x} \sim \mathbb{P}_r}[\log D(x)] + \nonumber 
\mathbb{E}_{\tilde{\mathbf{x}} \sim \mathbb{P}_g}[\log (1-D(\tilde{\mathbf{x}})], \label{cGAN_equation} 
\end{align}

\noindent where \(\mathbb{P}_r\) is the real/ground truth data distribution and \(\mathbb{P}_g\) is the model/generated data distribution defined by \(\tilde{\mathbf{x}} = G(\mathbf{I_{in}},\mathbf{M}, \mathbf{I^1_{pred}})\). Thus, our overall loss function for the refinement stage becomes
\begin{align} 
\mathcal{L}_{total}  = \lambda_1 \mathcal{L}_{\ell_1} + \lambda_2 \mathcal{L}_{adv}, 
%\label{objective}
\end{align}
where $\lambda _{1}=1, \lambda _{2}=0.1$. The weighted sum of these two loss functions compliments each other in the following way:
i) The GAN loss helps to improve the realism of the inpainted images, by fooling the discriminator.
ii) The \(\ell_1\) reconstruction loss serves as a regularization term for training GANs~\cite{yu2018generative}, helps in stabilizing GAN training, and encourages the generator to generate images from the modes that are close to the ground truth in an \(\ell_1\) sense.

\addtolength{\tabcolsep}{-3.0pt}    
\def\arraystretch{0.35}
\begin{figure*}[!t]
	\centering
	\begin{adjustbox}{width=1.0\textwidth}
		\scriptsize
		\begin{tabular} {ccccccccccccc}
			\small{Input} & 
			\small{PM~\cite{barnes2009patchmatch}} & \small{CE~\cite{pathak2016context}} & \small{CA~\cite{yu2018generative}} & 	\small{GI~\cite{yu2019free}} & \small{Ours} & \small{GT} &  
			\small{DFT~\cite{barnes2009patchmatch}} & \small{DFT~\cite{pathak2016context}} & \small{DFT~\cite{yu2018generative}} &
			\small{DFT~\cite{yu2019free}} & \small{DFT (Ours)} & \small{DFT (GT)} \\ \\
		
			\includegraphics[width=0.8in, height=0.8in]{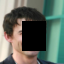} & 
			\includegraphics[width=0.8in, height=0.8in]{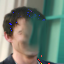} & 
			\includegraphics[width=0.8in, height=0.8in]{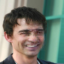} &			
			\includegraphics[width=0.8in, height=0.8in]{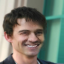} &			
			\includegraphics[width=0.8in, height=0.8in]{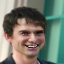} &			
			\includegraphics[width=0.8in, height=0.8in]{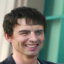} &		
			\includegraphics[width=0.8in, height=0.8in]{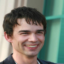} &	
			\includegraphics[width=0.8in, height=0.8in]{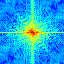} & 
			\includegraphics[width=0.8in, height=0.8in]{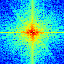} &
			\includegraphics[width=0.8in, height=0.8in]{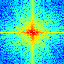} &
			\includegraphics[width=0.8in, height=0.8in]{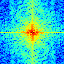} &
			\includegraphics[width=0.8in, height=0.8in]{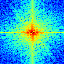} &
			\includegraphics[width=0.8in, height=0.8in]{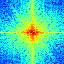} \\ \\
			PSNR/SSIM/$\ell_1$(\%) & 21.60/0.87/3.60 &30.86/0.98/0.92 &29.91/0.98/1.18 & 27.73/0.97/1.42 &\textbf{ 31.21/0.99/0.84} & & &&&&& \\ 
			
			\includegraphics[width=0.8in, height=0.8in]{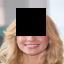} & 
			\includegraphics[width=0.8in, height=0.8in]{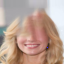} & 
			\includegraphics[width=0.8in, height=0.8in]{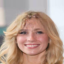} &
			\includegraphics[width=0.8in, height=0.8in]{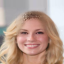} &
			\includegraphics[width=0.8in, height=0.8in]{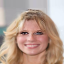} &
			\includegraphics[width=0.8in, height=0.8in]{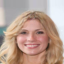} &
			\includegraphics[width=0.8in, height=0.8in]{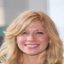} &
			\includegraphics[width=0.8in, height=0.8in]{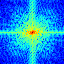} & 
			\includegraphics[width=0.8in, height=0.8in]{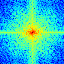} &
			\includegraphics[width=0.8in, height=0.8in]{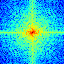} &
			\includegraphics[width=0.8in, height=0.8in]{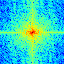} &
			\includegraphics[width=0.8in, height=0.8in]{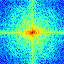} &
			\includegraphics[width=0.8in, height=0.8in]{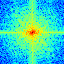} \\ \\
			PSNR/SSIM/$\ell_1$(\%) & 25.50/0.90/1.56 & 27.62/0.94/0.99 & 27.53/0.94/1.06 & 25.75/0.91/1.30 &\textbf{28.64/0.95/0.95}  & & &&&&& \\ \\
			
			\includegraphics[width=0.8in, height=0.8in]{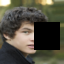} & 
			\includegraphics[width=0.8in, height=0.8in]{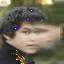} & 
			\includegraphics[width=0.8in, height=0.8in]{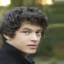} &
			\includegraphics[width=0.8in, height=0.8in]{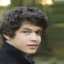} &
			\includegraphics[width=0.8in, height=0.8in]{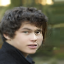} &
			\includegraphics[width=0.8in, height=0.8in]{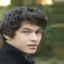} &
			\includegraphics[width=0.8in, height=0.8in]{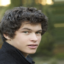} &
			\includegraphics[width=0.8in, height=0.8in]{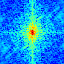} & 
			\includegraphics[width=0.8in, height=0.8in]{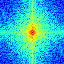} &
			\includegraphics[width=0.8in, height=0.8in]{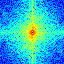} &
			\includegraphics[width=0.8in, height=0.8in]{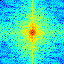} &
			\includegraphics[width=0.8in, height=0.8in]{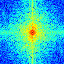} &
			\includegraphics[width=0.8in, height=0.8in]{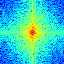} \\ \\
			PSNR/SSIM/$\ell_1$(\%) & 17.56/0.75/7.07 & \textbf{30.55/0.98/1.11} & 29.06/0.97/1.45 & 26.96/0.96/1.71 & 30.06/0.97/1.14 & & &&&&& \\ \\
		\end{tabular}
	\end{adjustbox}
	\vspace{1.5mm}
	\caption{Visual comparison of semantic feature completion results for different methods on CelebA dataset along with the DFT maps corresponding to different methods, our first stage output and GT image.}
	\label{fig:celeba_rbbox}	
\end{figure*}

\addtolength{\tabcolsep}{-0.8pt}    
\def\arraystretch{0.35}
\begin{figure*}[!t]
	\centering
	\begin{adjustbox}{width=\textwidth}
		\scriptsize
		\begin{tabular} {ccccccccccccc}
			\small{Input} & 
			\small{PM~\cite{barnes2009patchmatch}} & \small{CE~\cite{pathak2016context}} & \small{CA~\cite{yu2018generative}} & 	\small{GI~\cite{yu2019free}} & \small{Ours} & \small{GT} &  
			\small{DFT~\cite{barnes2009patchmatch}} & \small{DFT~\cite{pathak2016context}} & \small{DFT~\cite{yu2018generative}} &
			\small{DFT~\cite{yu2019free}} & \small{DFT (Ours)} & \small{DFT (GT)} \\ \\
						
			\includegraphics[width=0.8in, height=0.8in]{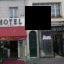} & 
			\includegraphics[width=0.8in, height=0.8in]{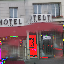} & 
			\includegraphics[width=0.8in, height=0.8in]{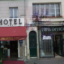} &			
			\includegraphics[width=0.8in, height=0.8in]{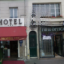} &			
			\includegraphics[width=0.8in, height=0.8in]{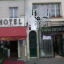} &			
			\includegraphics[width=0.8in, height=0.8in]{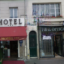} &
		    \includegraphics[width=0.8in, height=0.8in]{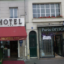} &		
		    	
			\includegraphics[width=0.8in, height=0.8in]{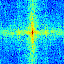} &
			\includegraphics[width=0.8in, height=0.8in]{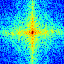} &
			\includegraphics[width=0.8in, height=0.8in]{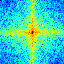} &
			\includegraphics[width=0.8in, height=0.8in]{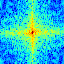} &
			\includegraphics[width=0.8in, height=0.8in]{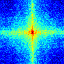} &
			\includegraphics[width=0.8in, height=0.8in]{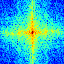} \\ \\
			
			 PSNR/SSIM/$\ell_1$(\%) & 17.32/0.74/10.24 & 24.94/0.94/1.97 & 25.49/0.95/1.94 & 24.74/0.95/2.28 & \textbf{28.98/0.98/1.26} & & &&&&& \\ \\
			
			\includegraphics[width=0.8in, height=0.8in]{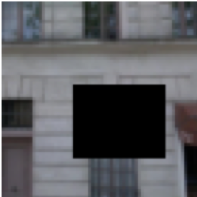} & 
			\includegraphics[width=0.8in, height=0.8in]{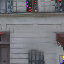} & 
			\includegraphics[width=0.8in, height=0.8in]{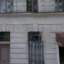} &
			\includegraphics[width=0.8in, height=0.8in]{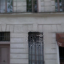} &
			\includegraphics[width=0.8in, height=0.8in]{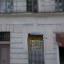} &			
			\includegraphics[width=0.8in, height=0.8in]{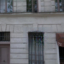} &			
			\includegraphics[width=0.8in, height=0.8in]{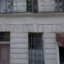} &
						
			\includegraphics[width=0.8in, height=0.8in]{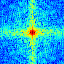} &
			\includegraphics[width=0.8in, height=0.8in]{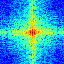} &
			\includegraphics[width=0.8in, height=0.8in]{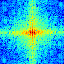} &
			\includegraphics[width=0.8in, height=0.8in]{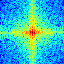} &
			\includegraphics[width=0.8in, height=0.8in]{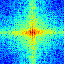} &
			\includegraphics[width=0.8in, height=0.8in]{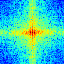}  \\ \\
			
			PSNR/SSIM/$\ell_1$(\%) & 18.15/0.55/6.88 & 30.01/0.95/1.09 & 29.29/0.95/1.17 & 25.62/0.90/1.83 &\textbf{30.11/0.96/1.08} & & &&&&&\\ \\
			
			\includegraphics[width=0.8in, height=0.8in]{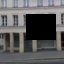} & 
			\includegraphics[width=0.8in, height=0.8in]{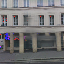} & 
			\includegraphics[width=0.8in, height=0.8in]{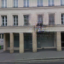} &			
			\includegraphics[width=0.8in, height=0.8in]{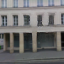} &			
			\includegraphics[width=0.8in, height=0.8in]{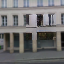} &			
			\includegraphics[width=0.8in, height=0.8in]{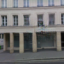} &		
			\includegraphics[width=0.8in, height=0.8in]{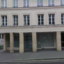} &	
			
			\includegraphics[width=0.8in, height=0.8in]{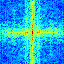} &
			\includegraphics[width=0.8in, height=0.8in]{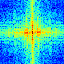} &	
			\includegraphics[width=0.8in, height=0.8in]{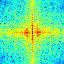} &
			\includegraphics[width=0.8in, height=0.8in]{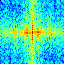} &
			\includegraphics[width=0.8in, height=0.8in]{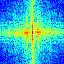} &
			\includegraphics[width=0.8in, height=0.8in]{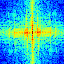} \\ \\
			
			PSNR/SSIM/$\ell_1$(\%) & 22.01/0.86/5.37 & 27.22/0.95/1.46 & 27.92/0.96/1.29 & 24.89/0.93/2.01 &\textbf{29.48/0.97/1.12}  & & &&&&&\\ 

		\end{tabular}
	\end{adjustbox}
	\vspace{1.5mm}
	\caption{Visual comparison of semantic feature completion results for different methods on Paris Streetview dataset along with the DFT maps corresponding to different methods, our first stage output and GT image.}
	\label{fig:PSV_rbbox}	
\end{figure*}

\section{\textbf{Implementation Details}}\label{implementation}
Our proposed model is implemented in PyTorch.~\footnote{Our code is available at \url{https://github.com/hiyaroy12/DFT_inpainting}.} In our experiments, we resize the image to 64$\times$64 and linearly scale the pixel values from range \([0, 256]\) to \([-1, 1]\). For the first stage, we initialize the weights by using He initialization~\cite{he2015delving} and use SGD optimizer with weight decay of 0.0001, the momentum of 0.9, and mini-batch size of 128. To train the first stage network we decayed the learning rate exponentially from $10^{-1}$ to $10^{-4}$ for 50 epochs. For the second stage, both our Generator G and Discriminator D are trained together using the following settings: i) G and D learning rate of $10^{-4}$, and $10^{-5}$ respectively, ii) optimized using Adam optimizer~\cite{kingma2014adam} with $\beta_{1} = 0.5$ and $\beta_{2} = 0.999$. In our experiments, we use a batch size of 14 and the training iterations of 100. Both stages are implemented on a TITAN Xp (12 GB) GPU.
\section{Experiments}
In this section, we evaluate the inpainting performance of our proposed method on three standard datasets: CelebFaces Attributes Dataset (CelebA)~\cite{liu2015deep}, Paris StreetView~\cite{doersch2012makes}, and Describable Texture Dataset (DTD)~\cite{cimpoi2014describing}. For our experiments, we use both regular and irregular masks. Regular masks refer to square masks having fixed size consisting of 25\% of total image pixels and are randomly located in the image. For irregular masks, during training, we use the masks from the work of Liu et al.~\cite{liu2018image}, where the irregular mask dataset contains the augmented versions of each mask~(0, 90, 180, 270 degrees rotated, horizontally reflected) and are divided based on the percentage of mask size on the image in increments of 10\% such as 0-10\%, 10-20\% etc.
%%%%%%%%%%%%%

\addtolength{\tabcolsep}{-0.9pt}    
\def\arraystretch{0.9}
\begin{figure*}[t]
	\centering
	\begin{adjustbox}{width=\textwidth}
		\scriptsize
		\begin{tabular} {ccccccccc}	
            & Input & GI~\cite{yu2019free} & Ours & GT & Input & GI~\cite{yu2019free} & Ours & GT \\
			\multirow{-6}{*}{\rotatebox{90}{10-20\% mask}} & 
			\includegraphics[width=0.8in, height=0.8in]{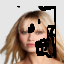} & 
			\includegraphics[width=0.8in, height=0.8in]{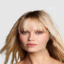} &
			\includegraphics[width=0.8in, height=0.8in]{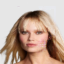} &
			\includegraphics[width=0.8in, height=0.8in]{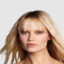} &
			\includegraphics[width=0.8in, height=0.8in]{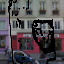} & 
			\includegraphics[width=0.8in, height=0.8in]{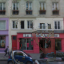} & 
			\includegraphics[width=0.8in, height=0.8in]{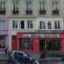} & 
			\includegraphics[width=0.8in, height=0.8in]{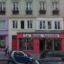} \\
			& \scriptsize{PSNR/SSIM/$\ell_1$(\%)} & 31.92/0.993/0.46 & \textbf{33.12/0.99/0.37 }& & \scriptsize{PSNR/SSIM/$\ell_1$(\%)}& 
			28.99/0.96/1.56 & \textbf{32.23/0.99/1.02} &  \\
			\multirow{-6}{*}{\rotatebox{90}{20-30\% mask}} & 
			\includegraphics[width=0.8in, height=0.8in]{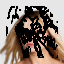} &
			\includegraphics[width=0.8in, height=0.8in]{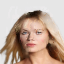} &
			\includegraphics[width=0.8in, height=0.8in]{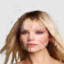} &
			\includegraphics[width=0.8in, height=0.8in]{figures/irregular/clean/162781} &
			\includegraphics[width=0.8in, height=0.8in]{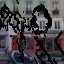} &
			\includegraphics[width=0.8in, height=0.8in]{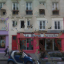} &
			\includegraphics[width=0.8in, height=0.8in]{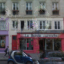} &
			\includegraphics[width=0.8in, height=0.8in]{figures/irregular/clean/001_im}\\
			& \scriptsize{PSNR/SSIM/$\ell_1$(\%)} & 24.99/0.96/1.46 & \textbf{27.46/0.98/1.21 }& & \scriptsize{PSNR/SSIM/$\ell_1$(\%)}&  
			26.93/0.95/2.38 & \textbf{28.69/0.97/1.82} \\
			\multirow{-6}{*}{\rotatebox{90}{30-40\% mask}} & 
			\includegraphics[width=0.8in, height=0.8in]{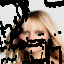} &
			\includegraphics[width=0.8in, height=0.8in]{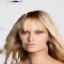} &
			\includegraphics[width=0.8in, height=0.8in]{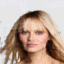} & 
			\includegraphics[width=0.8in, height=0.8in]{figures/irregular/clean/162781} &
			\includegraphics[width=0.8in, height=0.8in]{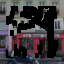} &
			\includegraphics[width=0.8in, height=0.8in]{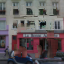} &
			\includegraphics[width=0.8in, height=0.8in]{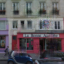} &
			\includegraphics[width=0.8in, height=0.8in]{figures/irregular/clean/001_im}\\
			& \scriptsize{PSNR/SSIM/$\ell_1$(\%)} & 24.52/0.97/1.62  & \textbf{27.09/0.99/1.29}  && \scriptsize{PSNR/SSIM/$\ell_1$(\%)}& 
			23.50/0.89/3.94 & \textbf{26.86/0.95/2.58 } & \\
			\multirow{-6}{*}{\rotatebox{90}{40-50\% mask}} &
			\includegraphics[width=0.8in, height=0.8in]{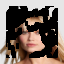} &
			\includegraphics[width=0.8in, height=0.8in]{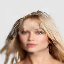} &
			\includegraphics[width=0.8in, height=0.8in]{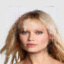} & 
			\includegraphics[width=0.8in, height=0.8in]{figures/irregular/clean/162781} &
			\includegraphics[width=0.8in, height=0.8in]{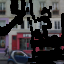} &
			\includegraphics[width=0.8in, height=0.8in]{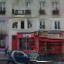} &
			\includegraphics[width=0.8in, height=0.8in]{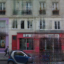} & 
			\includegraphics[width=0.8in, height=0.8in]{figures/irregular/clean/001_im}\\
			& \scriptsize{PSNR/SSIM/$\ell_1$(\%)} & 20.31/0.89/2.66 &\textbf{ 25.86/0.97/1.59 }& &  \scriptsize{PSNR/SSIM/$\ell_1$(\%)}& 
			22.45/0.90/5.04 & \textbf{25.54/0.94/3.43} \\
			\multirow{-6}{*}{\rotatebox{90}{50-60\% mask}} & 
			\includegraphics[width=0.8in, height=0.8in]{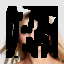} &
			\includegraphics[width=0.8in, height=0.8in]{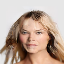} &
			\includegraphics[width=0.8in, height=0.8in]{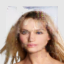} &
			\includegraphics[width=0.8in, height=0.8in]{figures/irregular/clean/162781} &
			\includegraphics[width=0.8in, height=0.8in]{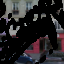} &
			\includegraphics[width=0.8in, height=0.8in]{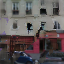} &
			\includegraphics[width=0.8in, height=0.8in]{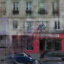} &
			\includegraphics[width=0.8in, height=0.8in]{figures/irregular/clean/001_im} \\
			& \scriptsize{PSNR/SSIM/$\ell_1$(\%)} & 21.17/0.92/3.04   &\textbf{ 22.36/0.94/2.74 }&& \scriptsize{PSNR/SSIM/$\ell_1$(\%)}& 
			19.30/0.76/8.30  & \textbf{22.04/0.85/5.98}  \\
		\end{tabular}
	\end{adjustbox}
	\vspace{1.5mm}
	\caption{Visual comparison of semantic feature completion results for irregular masks on CelebA and Paris StreetView dataset.}
	\label{fig:irregular}	
\end{figure*}

\subsection{Qualitative Evaluation}
Figures~\ref{fig:celeba_rbbox} and \ref{fig:PSV_rbbox} compare the inpainting results of our method with previous image inpainting methods: PatchMatch~(PM)~\cite{barnes2009patchmatch}, Context Encoder~(CE)~\cite{pathak2016context}, Contextual Attention~(CA)~\cite{yu2018generative}, and Generative Inpainting~(GI)~\cite{yu2019free}, for regular masks on CelebA and Paris StreetView datasets. The last six columns of these figures demonstrate the magnitude spectrum of the DFT map obtained from different methods~\cite{barnes2009patchmatch,pathak2016context,yu2018generative,yu2019free}, our method (first stage reconstruction) and the ground truth image. We can see that previous methods (PM) copy incorrect patches in the missing regions, whereas others (CE, CA, GI) sometimes fail to achieve plausible results and generate distinct artifacts. However, our method can restore the missing regions with sharp structural details, minimal blurriness, and hardly any ``checkerboard'' artifacts. Moreover, the inpainting results using our method look the most similar to the ground truth images. We conjecture that in the presence of frequency domain information, the network efficiently learns the high-frequency details, which enables it to preserve the structural details in the restored image. This can be confirmed from the DFT maps where we see that our deconvolution network learns to predict the missing region in such a way that the DFT map of our first stage reconstruction looks similar to that of the ground truth image. Later the refinement network uses this frequency domain information to produce better inpainting results. 

We also show the performance of our proposed method on CelebA and Paris StreetView dataset for irregular masks. Figure~\ref{fig:irregular} shows the inpainting results using Generative Inpainting~(GI)~\cite{yu2019free} and our proposed method for different percentage~(10-50\%) of mask size. Our method can generate photo-realistic images having similar texture and structures as the original clean images even when a large region (50-60\%) of the image is missing. 
\subsection{Quantitative Evaluation}
We report the quantitative performance of our method in terms of the following metrics i) peak-signal-to-noise ratio~(PSNR); ii) structural similarity index~(SSIM)~\cite{wang2004image} and iii) mean absolute error~(MAE). Table~\ref{table:metric} demonstrates the comparison in metric values on the CelebA, Paris StreetView, and DTD dataset for the state-of-the-art inpainting methods and our method. Our method outperforms previous methods in terms of these metrics on both regular and irregular masks. This proves the effectiveness of using frequency domain information. Note that, we obtain the metrics for Context Encoder~\cite{pathak2016context} by using the \(\ell_1\) and adversarial loss in our network settings. 

\addtolength{\tabcolsep}{2.0pt}    
\begin{table*}[t]
	\def\arraystretch{1.3}
	\centering
	\caption{Quantitative results on CelebA~\cite{liu2015deep}, Paris Streetview~\cite{doersch2012makes}, and DTD texture dataset~\cite{cimpoi2014describing} for different inpainting models: PatchMatch~(PM)~\cite{barnes2009patchmatch}, Context Encoder~(CE)~\cite{pathak2016context}, Contextual Attention~(CA)~\cite{yu2018generative}, Generative Inpainting~(GI)~\cite{yu2019free}, and Ours. The best results for each row is shown in bold. $^-$Lower is better. $^+$Higher is better.} 
	\begin{adjustbox}{width=\textwidth}
		\begin{tabular}{cc*{16}{>{\centering\arraybackslash}p{.09\linewidth}}}			
			& & \multicolumn{5}{c}{\large{CelebA dataset}} & \multicolumn{5}{c}{\large{Paris Streetview dataset}} & \multicolumn{5}{c}{\large{DTD texture dataset}} \\  \cmidrule(r){3-7} \cmidrule(r){8-12} \cmidrule(r){13-17}
			\multicolumn{2}{r}{\textbf{Mask}} 
			& \small{PM~\cite{barnes2009patchmatch}} & \small{CE~\cite{pathak2016context}} & \small{CA~\cite{yu2018generative}} & \small{GI~\cite{yu2019free}} & \small{Ours} 
			& \small{PM~\cite{barnes2009patchmatch}} & \small{CE~\cite{pathak2016context}} & \small{CA~\cite{yu2018generative}} & \small{GI~\cite{yu2019free}} & \small{Ours} 
			& \small{PM~\cite{barnes2009patchmatch}} & \small{CE~\cite{pathak2016context}} & \small{CA~\cite{yu2018generative}} & \small{GI~\cite{yu2019free}}  & \small{Ours} \\ \hhline{*{16}{=}=}
			
			\multirow{5}{*}{\rotatebox{90}{PSNR$^{+}$}}
			& \small{10-20\%} & 15.78 & 32.49 & 29.81  & 30.65 & \textbf{32.69} & 22.03 & 31.59 & 30.68 & 30.42 & \textbf{32.34} & 22.43 & 29.28 & 28.43 & 29.29  & \textbf{29.89}  \\ 
			& \small{20-30\%} & 15.09 & 29.62 & 27.06  & 27.22 & \textbf{29.78} & 20.42 & 28.69 & 27.40 & 27.09 & \textbf{29.25} & 21.11 & 27.02 & 25.73 & 26.34  & \textbf{27.38} \\ 
			& \small{30-40\%} & 14.42 &  27.31 & 24.77  & 24.83 &  \textbf{27.49} & 19.36 & 27.02 & 25.42 & 24.95 & \textbf{27.33} & 20.12 & 25.33 & 23.76 & 24.41 & \textbf{25.65} \\ 
			& \small{40-50\%} & 13.63 & 25.10 & 23.03& 22.86 & \textbf{25.27} & 18.52 & 25.09 & 23.99 & 23.23 & \textbf{25.13} & 19.26 & 23.89 & 22.35 &  22.75 & \textbf{23.95}  \\ 
			& \small{Regular}   & 14.96 & \textbf{28.17} & 27.86 & 26.06 &  28.13 & 19.23 & 27.32  & 28.29  & 25.12 & \textbf{28.42} & 14.75 & 27.33 & 27.26 &  25.73  & \textbf{27.49} \\ \hhline{*{16}{-}-}	
			\multirow{5}{*}{\rotatebox{90}{SSIM$^{+}$}}
			
			& \small{10-20\%} & 0.632 & 0.991 & 0.986 & 0.987  & \textbf{0.992} & 0.766 & 0.978 & 0.972 & 0.969  & \textbf{0.981} & 0.704 & 0.933 & 0.922  & 0.935  & \textbf{0.942} \\ 
			& \small{20-30\%} & 0.579 & 0.983 & 0.971 & 0.971 & \textbf{0.984} & 0.692 & 0.958 & 0.945 & 0.936  & \textbf{0.963} & 0.634 & 0.890 & 0.861  & 0.872  & \textbf{0.901} \\
			& \small{30-40\%} & 0.513 & 0.972 & 0.953  & 0.952 & \textbf{0.973} & 0.613 & 0.938  & 0.912  & 0.896  & \textbf{0.942} & 0.563 & 0.841 & 0.793  & 0.804  & \textbf{0.854} \\ 
			& \small{40-50\%} & 0.421 & 0.954 & 0.930 & 0.927 & \textbf{0.956}  & 0.515 & 0.904 & 0.873 & 0.850 & \textbf{0.910} & 0.475 & 0.773 & 0.717  & 0.714  & \textbf{0.785} \\ 
			& \small{Regular}  & 0.571 & 0.970 & 0.968 & 0.953 &  \textbf{0.971} & 0.659 & 0.923 & 0.934  &  0.880  &  \textbf{0.936} & 0.149 & 0.876 & 0.869  & 0.833  & \textbf{0.879} \\ \hhline{*{16}{-}-}
			\multirow{5}{*}{\rotatebox{90}{$\ell_1$ (\%)$^{-}$}}
			
			& \small{10-20\%} & 13.14 & 0.84 & 1.37 &  1.21 & \textbf{0.82} & 6.15 & 1.09 & 1.40   & 1.44  & \textbf{0.97} & 7.87 & 1.87 & 1.92  & 1.81 & \textbf{1.67} \\ 
			& \small{20-30\%} & 14.58 & 1.41 &  2.24 & 2.07 & \textbf{1.39}  & 7.78 & 1.93 & 2.45   &  2.52 & \textbf{1.78} & 8.85 & 2.85 & 3.02  & 2.93  & \textbf{2.62} \\ 
			& \small{30-40\%} & 16.07 & 2.13 & 3.28 & 3.09 & \textbf{2.09}  & 9.39 & 2.70 & 3.43   & 3.66  & \textbf{2.57} & 9.76 & 3.82 & 4.20  &  4.11 & \textbf{3.58} \\
			& \small{40-50\%} & 17.89 & 3.13 & 4.40 & 4.22 & \textbf{3.08 } & 10.8 & 3.75 & 4.40   & 4.79  & \textbf{3.58} & 10.70 & 4.94 & 5.40  & 5.43  & \textbf{4.74} \\ 
			& \small{Regular} &  13.67 & 1.55 & 1.76 & 2.12  & \textbf{1.55} & 9.04 & 1.97 & 1.93   & 2.76  & \textbf{1.77}  & 17.60 & 2.12 & 2.40  & 2.74 & \textbf{2.05} \\ \hline
		\end{tabular}
	\end{adjustbox}
	\vspace{1.5mm}
	\label{table:metric}
\end{table*}

\begin{figure*}[tb]
	\centering
	\includegraphics[width=0.95\linewidth]{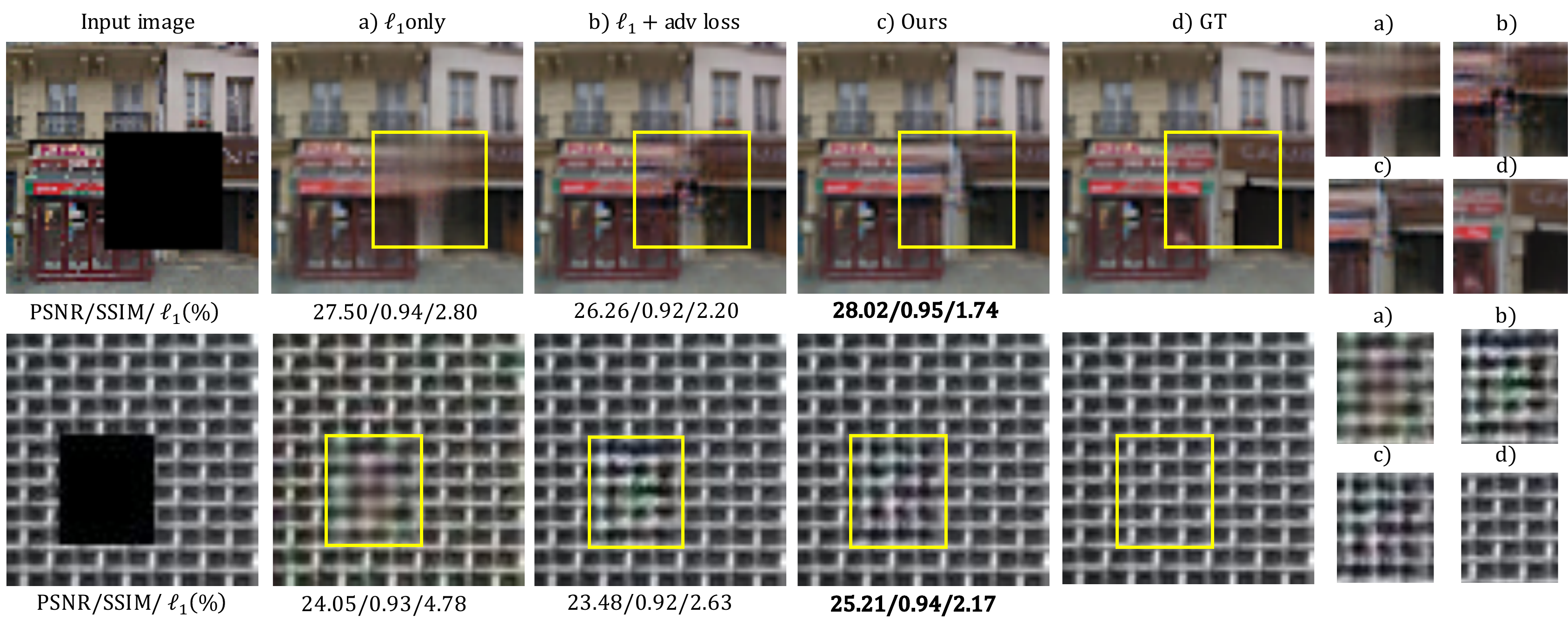}
	\vspace{1mm}
	\caption{Visual results on Paris StreetView dataset (first row) and DTD (second row) showing the effect of different components in our model on the input incomplete images (first column), a) results using standard $\ell_1$ loss, b) results using $\ell_1$ + adversarial loss, c) results of our model trained using $\ell_1$ + adversarial loss (with DFT component), and d) GT image.}\label{fig:ablation}
\end{figure*}
\subsection{Ablation Study}
We perform an ablation study to investigate the role of our frequency deconvolution network and to analyze the effect of different loss components used to train our model. Figure~\ref{fig:ablation} shows the inpainting results using only \(\ell_1\) loss, \(\ell_1\) with adversarial loss and our proposed method of incorporating frequency domain information (DFT component). We can see blurry reconstructions in Figure~\ref{fig:ablation}a) when we use only \(\ell_1\) loss in the spatial domain. However, inpainting performance improves to a certain extent if we add the adversarial loss component. Nevertheless, in Figure~\ref{fig:ablation}b) we can still find structural and blurry artifacts on the reconstructions. Figure~\ref{fig:ablation}c) demonstrates the inpainting results of our proposed method of training the model using both frequency and spatial components. We can see in that using our method the model can perform significantly better by restoring fine structural details. Therefore, we can conclude that training the model along with frequency-domain information certainly helps the network to learn high-frequency components and restore the missing region with better reconstruction quality.
\section{Conclusion}
We presented a frequency-based image inpainting algorithm that enables the network to use both frequency and spatial information to predict the missing region of an image. Our model first learned the global context using frequency domain information and selectively reconstructed the high-frequency components. Then it used the spatial domain information as a guidance to match the color distribution of the true image and fine-tuned the details and structures obtained in the first stage, leading to better inpainting results. Experimental results showed that our method could achieve results better than state-of-the-art performances on challenging datasets by generating sharper details and perceptually realistic inpainting results. Based on our empirical results, we believe that methods using both frequency and spatial information should gain dominance because of their superior performance. In the future, we want to extend this work to using other kinds of frequency domain transformations e.g. Discrete Cosine Transform and solve other kinds of image restoration tasks e.g. image denoising.

\section{Acknowledgments}
This paper is partially financially supported by JSPS KAKENHI Grant (JP19K22863).
\bibliography{report}  
\bibliographystyle{spiejour} 

\vspace{2ex}\noindent\textbf{Hiya Roy} is a Ph.D. candidate in the Electrical Engineering and Information Systems department at the University of Tokyo. She received the B.E. degree in electrical engineering from Jadavpur University, India in 2012. She received the M.S. degree in electrical engineering and information systems from the University of Tokyo, Japan in 2017. Her research interests are computer vision, machine learning, deep learning and planetary sciences. She is a MEXT scholar from Sept 2015 to Aug 2020. She is a member of IEEE.

\vspace{2ex}\noindent\textbf{Subhajit Chaudhury} is a Research Scientist at IBM Research AI, Tokyo. Concurrently, he is a Ph.D. candidate at the Department of Information and Communication Engineering, The University of Tokyo. He received the B.E. degree in electrical engineering from Jadavpur University, India in 2012. He received the M.Tech. degree in Electrical Engineering from Indian Institute of Technology Bombay in 2014. He worked as a researcher at NEC Research Laboratories, Japan from Oct 2014 to Mar 2017. His current research interests include reinforcement learning and computer vision. He is a member of IEEE, and ACM.

\vspace{2ex}\noindent\textbf{Toshihiko Yamasaki} is currently an Associate Professor at Department of Information and Communication Engineering, Graduate School of Information Science and Technology, The University of Tokyo. He received the B.S. degree in electronic engineering, the M.S. degree in information and communication engineering, and the Ph.D. degree from The University of Tokyo in 1999, 2001, and 2004, respectively. He was a JSPS Fellow for Research Abroad and a visiting scientist at Cornell University from Feb. 2011 to Feb. 2013. His current research interests include attractiveness computing based on multimedia big data analysis, pattern recognition, machine learning, and so on. His publication includes three book chapters, more than 60 journal papers, more than 160 international conference papers. Dr. Yamasaki is a member of IEEE, ACM, IEICE, ITE, IPSJ.

\vspace{2ex}\noindent\textbf{Tatsuaki Hashimoto} is a professor of ISAS, Japan Aerospace Exploration Agency and also Graduate school of engineering, the University of Tokyo. He received Ph.D of Electrical Engineering from The University of Tokyo in 1990. Form April, 1990, he has been working for the Institute of Space and Astronautical Science (ISAS), Currently, he  He has been working for research and development of spacecraft guidance, navigation, and control system. He stayed at Jet Propulsion Laboratory, NASA as a visiting scientist in 2000, His research interest includes image processing for spacecraft navigation and lunar surface exploration. He is a member of AIAA, IAA, IEEJ, SICE, etc.
\end{document}